\documentclass{article}

\usepackage{arxiv}
\usepackage{array}
\usepackage[utf8]{inputenc} 
\usepackage[T1]{fontenc}    
\usepackage{hyperref}       
\usepackage{url}            
\usepackage{booktabs}       
\usepackage{amsfonts}       
\usepackage{nicefrac}       
\usepackage{microtype}      
\usepackage{lipsum}		
\usepackage{graphicx}
\usepackage{amssymb}
\usepackage{doi}
\usepackage{booktabs}

\title{Reinforced Large Language Model is a formal theorem prover}

\author{ Zhiling Luo \\
	Alibaba Group\\
	\texttt{godot.lzl@alibaba-inc.com}
}

\begin{document}
\maketitle

\begin{abstract}
To take advantage of Large Language Model in theorem formalization and proof, we propose a reinforcement learning framework to iteratively optimize the pretrained LLM by rolling out next tactics and comparing them with the expected ones. The experiment results show that it helps to achieve a higher accuracy compared with directly fine-tuned LLM.
\end{abstract}

\section{Introduction}
Theorem formalization and proof are fundamental processes in mathematics and computer science that involve expressing mathematical statements in a precise, unambiguous language and rigorously demonstrating their truth or validity. Formalization typically begins by translating an informal mathematical theorem into a formal logical system, where every term, assumption, and conclusion is explicitly defined using axioms, rules of inference, and symbolic notation. This ensures that the statement is free from ambiguity and can be manipulated systematically. Once formalized, the proof process involves constructing a sequence of logically valid steps, starting from agreed-upon axioms or previously proven theorems, to establish the desired result. Proofs can range from simple direct arguments to complex constructions involving advanced techniques like induction, contradiction, or model theory. 

In recent years, computer-assisted proof systems, such as Coq\cite{coq}, Isabelle\cite{isabelle}, and Lean\cite{lean}, have emerged as powerful tools for verifying formalized theorems with unprecedented precision, enabling mathematicians to tackle increasingly intricate problems while minimizing human error. Together, theorem formalization and proof serve as cornerstones of reliable knowledge, ensuring that mathematical discoveries are both robust and enduring. 

Large Language Models (LLMs) are advanced artificial intelligence systems designed to understand and generate human-like text across a wide range of topics. These models are trained on vast amounts of data, enabling them to perform tasks such as answering questions, writing essays, creating code, and even engaging in meaningful conversations. By leveraging deep learning techniques, LLMs can capture complex patterns in language, making them highly versatile tools for various applications, from education and research to business and entertainment. Their ability to adapt to different contexts and provide coherent, contextually relevant responses has made LLMs a transformative force in the field of natural language processing. 

The success of LLM in solving NLP tasks arouse our interests to apply it in theorem formalization and proving. Recent researches InternLM \cite{InternLM} and lean-star \cite{lean-star} wraps LLM as an agent to complete the proof. And the LLMs are fine-tuned from a general base model. Its performance  is hampered by the high abstraction of formal language. e.g. Lean 4. Specifically, the most theorem related data collected and trained in LLM are informal. Some are presented in the latex style and others are normal natural languages. However, the formal statement are different and involving more unicode symbols. A comparison is illustrated by Tab. \ref{tab:informal-formal-theorems}. It limits the LLM's ability to understand and reason about formal theorems. To alleviate this dilemma, we study the research question: How to post-train the pretrained LLM as a better formal theorem prover?

In this paper, we propose a reinforced learning framework to iteratively update the LLM by rolling out next tactics and comparing them with the expected ones. The experiment on miniF2F\cite{minif2f} show the advantages comparing with directly fine-tune LLMs.
The framework and trained LLM based on Qwen0.5B \cite{qwen} is open-sourced to the community at Github
\url{https://github.com/zhilingluo/theorem_prover} .
\begin{table}[h!]
    \centering
    \begin{tabular}{>{\raggedright\arraybackslash}p{3cm} >{\raggedright\arraybackslash}p{5cm} >{\raggedright\arraybackslash}p{7cm}}
        \toprule
        \textbf{Description} & \textbf{Informal Statement} & \textbf{Formal Statement in Lean} \\
        \midrule
        Pythagorean Theorem & In a right triangle, the square of the hypotenuse is equal to the sum of the squares of the other two sides.
        &
        \texttt{theorem pythagorean\_theorem (a b c : $\mathbb{R}$) (h : $a^2$ + $b^2$ = $c^2$) : right\_triangle a b c := ...}
        \\
        \midrule
        Associativity of Addition & For all real numbers $a$, $b$, and $c$, $(a + b) + c = a + (b + c)$.
        &
        \texttt{theorem add\_assoc (a b c : $\mathbb{R}$) : (a + b) + c = a + (b + c) := ...}
        \\
        \midrule
        Existence of Identity for Addition & There exists a real number $0$ such that for all $a$, $a + 0 = a$.
        &
        \texttt{theorem zero\_add (a : $\mathbb{R}$) : 0 + a = a := ...}
        \\
        \bottomrule
    \end{tabular}
    \caption{Comparison of Informal and Formal Theorems in Lean}
    \label{tab:informal-formal-theorems}
\end{table}
\section{Framework}
\begin{figure}
    \centering
    \includegraphics[width=1\linewidth]{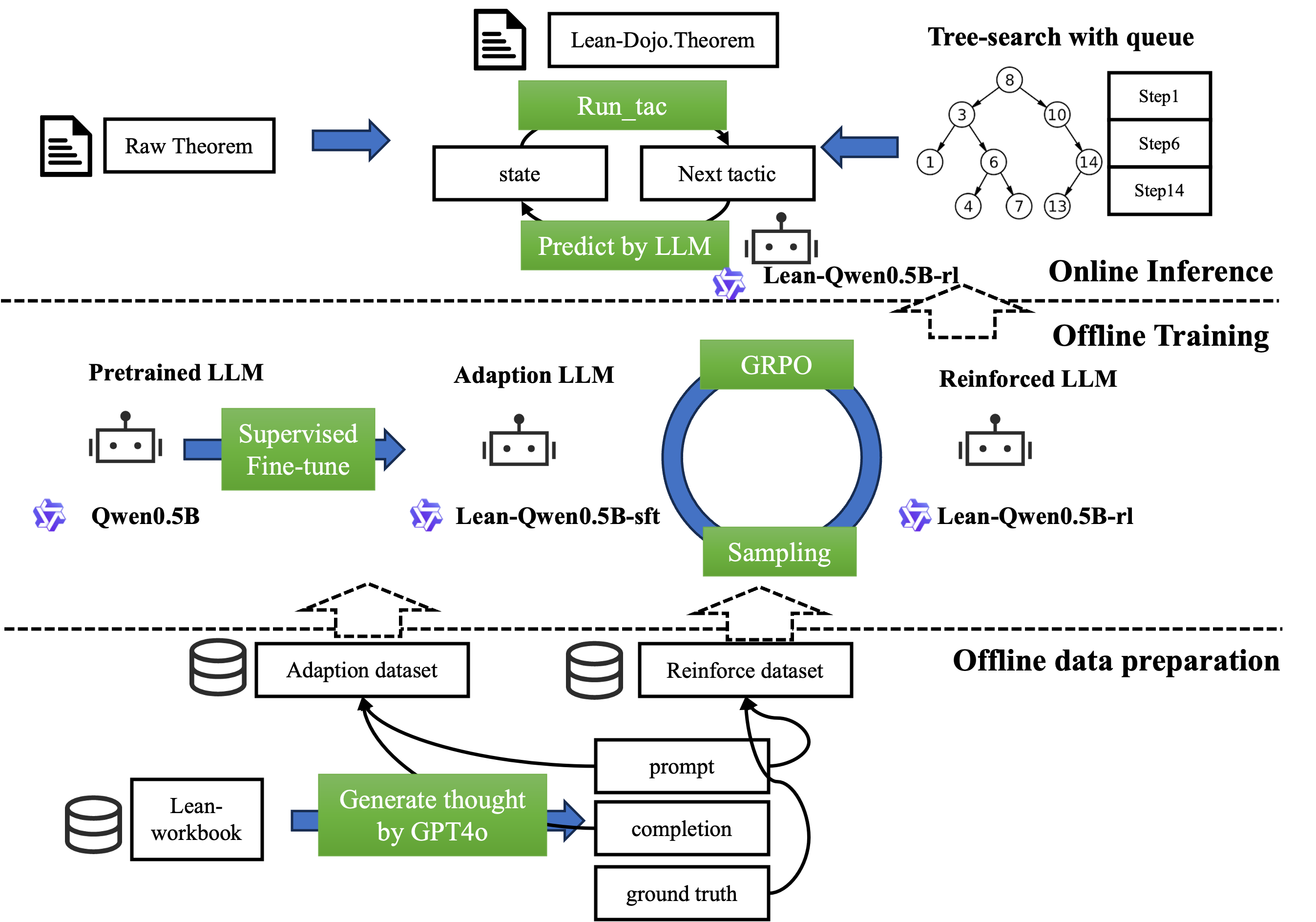}
    \caption{The framework includes offline data preparation, model training and online inference.}
    \label{fig:framework}
\end{figure}

In this section, we present the framework to train an reinforced LLM as theorem prover, as shown in Fig. \ref{fig:framework}.  This framework is designed to train an LLM for theorem proving within the Lean proof assistant environment. It consists of three parts: data preparation, model training and online inference. 
\subsection{Data Preparation}
The data preparation leverage Lean-workbook of InternLM as the raw dataset. Our key idea is to enhance the dataset by adding useful thought as a supervisory signal of CoT in training. We involve GPT4o by prompting "Read the following Lean4 theorem proving process, analyze the proposition to be proved and the available conditions, and provide the next tactic. Note that a reference next tactic is provided; do not assume prior knowledge of this reference". At the same time, we provide current state and next tactic of groundtruth to GPT4o. It constructs a paragraph of thought for each sample.
After that we prepare three items for each sample: prompt, completion and groundtruth. the prompt part is in the conversation format, like "\{'role':system','content':system\_prompt\}\{'role':'user','content':user\_prompt\}" where the system\_prompt is "You need to complete the proof in Lean4. Please think carefully and provide the next step based on the current state. The format should be:\textbackslash n<think>Your thought process</think>\textbackslash n<answer>```lean \textbackslash n Your strategy\textbackslash n```</answer>"
The completion part contains the thought and ground-truth tactic. At last the ground-truth part is exactly the next tactic.

We introduce two kind of dataset in our framework, one is the adaption dataset and another is the reinforce dataset. The former one contains only pair of prompt and completion. And the latter contains the pair of prompt and ground-truth.
\subsection{Training}
In the training part, we starts from a pretrained LLM, say Qwen0.5B \cite{qwen}. The entire training contains two phases, the first phase is adaption, and the second phase is reinforcement learning.
In the adaption phase, the pretrained LLM is supervised by the adaption dataset and updated as the adaption LLM. In our setup, the adaption LLM is named as Lean-Qwen0.5-sft. Recall that the adaption dataset contains the prepared thought as CoT, we need this phase to awake the basic CoT capacity of pretrained-LLM. 
The reinforcement phase is an iteration of sampling and optimization. In the sampling step, The LLM predict a group of completions as the candidates. In the optimization step, the reward is calculated by comparing the candidate with the ground-truth. After that the loss of reward weighted by advantage is calculated, also with the gradient w.r.t. parameters. We leverage GRPO as the optimizer in experiment. These two steps are iterated several times. In each iteration the current model, called policy model, samples candidates and is updated. After enough iteration, the model is derived as the reinforced LLM. In our setup, it is named as Lean-Qwen0.5B-rl. 
\subsection{Inference}
This part inputs a theorem, which is processed via the Lean-Dojo \cite{lean-dojo} as the dojo.Theorem object. We follow the framework of InternLM to interact LLM with theorem through a queue enhanced tree search. It queues steps to explore from search tree. The predicted tactic from LLM is run by run\_tac and the state is collected. It  has three situations:
1). The state is proof finish. It means current theorem is proved.
2). The state is a novel state, It adds a new node in the tree to explore.
3). The state is a grammar error. It means the tactic is wrong.
The queue is the breath-first search on the tree.

\section{Experiment}

\begin{table}[t]
    \centering
    \begin{tabular}{cccc}
        \toprule
         & \#Positive sample on MiniF2F & Acc on MiniF2F & Acc on Trainset \\
        \midrule
        Lean-qwen0.5B-sft & 11 & 36\% & 49\% \\
        Lean-qwen0.5B-rl & 13 & 43\% & 61\% \\
        \bottomrule
    \end{tabular}
    \caption{Acc comparison of sft and rl version on miniF2F and trainset}
    \label{tab:expmain}
\end{table}
The experiment is conducted on miniF2F containing 30 samples.  The raw training data source is Lean-workbook, which contains 25k samples. In the data preparation, we use GPT4o, whose version is chatgpt-4o-latest in Feb. 2025. The base pretrained LLM is Qwen2.5-0.5B.  the supervised fine-tune and grpo are implemented on TRL. The training process is conducted on an 4*A100 GPU. The accelerate lib is leveraged for data parallelization. Both adaption and reinforce phase take 1 day. In the inference part, Lean-Dojo is used and modified a little for LLM-Lean interaction. The tree-search is developed from the source code of InternLM. The trained LLM is deployed on Ollama with CPU setup. The inference is executed on a Macbook of Apple M2 Pro. The entire inference takes 1 hour. We use an additional format reward like \cite{deepseek}.

\begin{figure}
    \centering
    \includegraphics[width=1\linewidth]{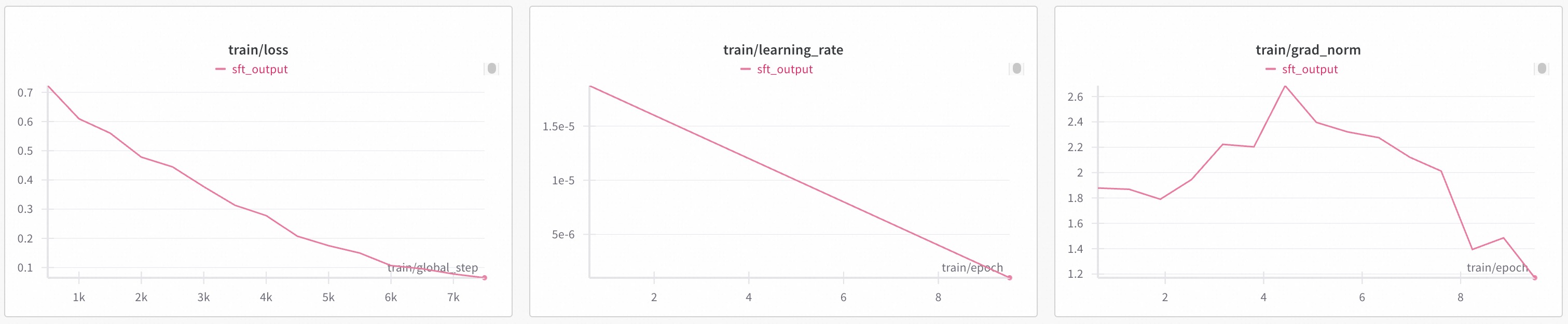}

    \caption{Train process records of adaption phase}
    \label{fig:train_sft}
\end{figure}

\begin{figure}
    \centering
    \includegraphics[width=1\linewidth]{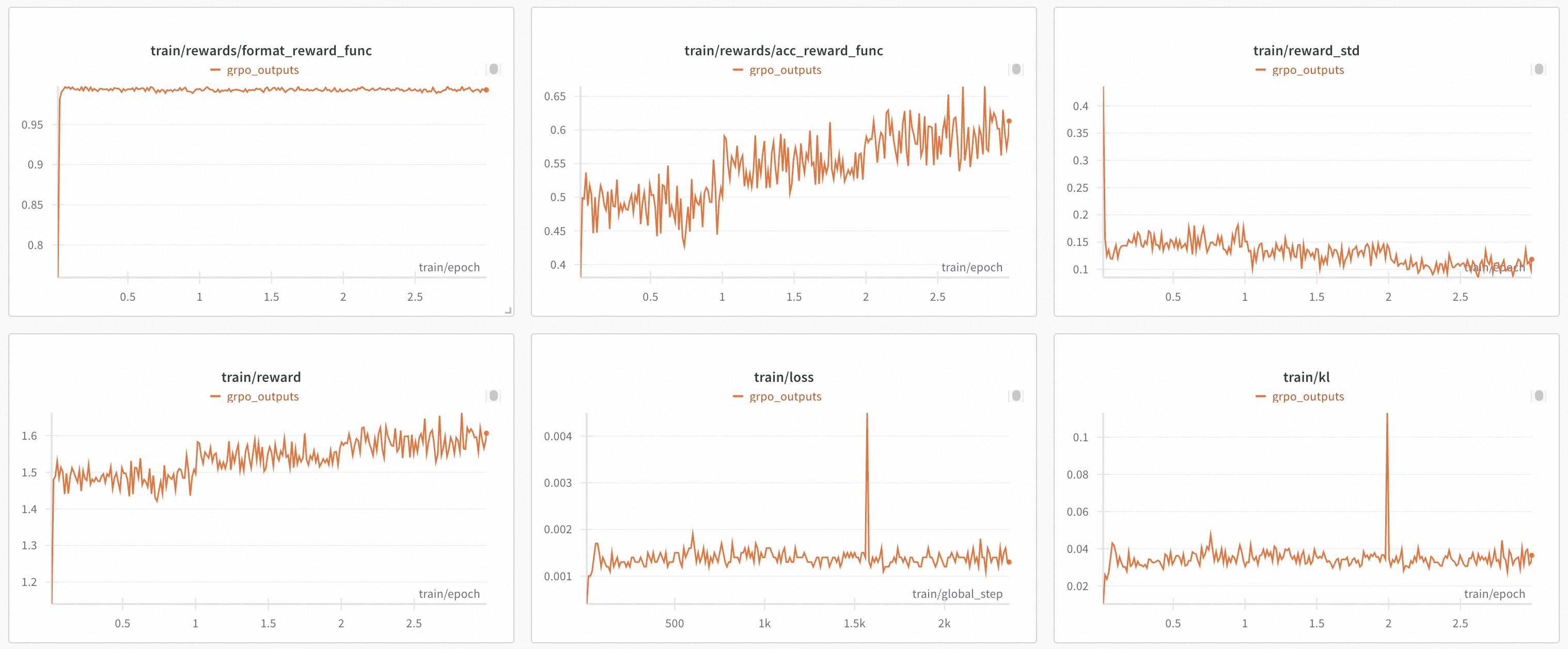}
    \caption{Train process records of RL phase}
    \label{fig:train_grpo}
\end{figure}

The train records of adaption and reinforce phases are presented in Fig. \ref{fig:train_sft} and Fig.\ref{fig:train_grpo}. We observe that :
1) the adaption training is close to converge and the loss comes to 0.1.
2) the reinforce training promotes the format reward very quick. And the accuracy reward curve is step-like. Specifically, it is stable in an epoch and increases a lot when comes a new epoch. 

The accuracy is reported in Tab. \ref{tab:expmain}, where our reinforced model outperforms the baseline model, which directly uses supervised finetune.

\section{Conclusion}
This paper studies a two-phase reinforcement learning framework to train a LLM as a formal theorem prover. The experiment results show that it helps to achieve a higher accuracy compared with supervised finetune. 



\bibliographystyle{plain}
\bibliography{main}  

\end{document}